\documentclass[conference]{IEEEtran}
\IEEEoverridecommandlockouts

\usepackage{cite}
\usepackage{amsmath,amssymb,amsfonts}
\usepackage{algorithmic}
\usepackage{graphicx}
\usepackage{textcomp}
\usepackage{bookmark}
\usepackage{hyperref}
\usepackage{CJKutf8}
\usepackage{xcolor}
\def\BibTeX{{\rm B\kern-.05em{\sc i\kern-.025em b}\kern-.08em
    T\kern-.1667em\lower.7ex\hbox{E}\kern-.125emX}}
\begin{document}

\title{Knowledge Management for Automobile Failure Analysis Using Graph RAG}

\author{\IEEEauthorblockN{1\textsuperscript{st} Yuta Ojima}
\IEEEauthorblockA{\textit{Department of System Innovation} \\
\textit{School of Engineering} \\
\textit{The University of Tokyo}\\
Tokyo, Japan \\
{y\_ojima}@m.sys.t.u-tokyo.ac.jp}
\and
\IEEEauthorblockN{2\textsuperscript{nd} Hiroki Sakaji}
\IEEEauthorblockA{\textit{Faculty of Information Science} \\
\textit{and Technology} \\
\textit{Hokkaido University}\\
Hokkaido, Japan \\
{sakaji}@ist.hokudai.ac.jp}
\and
\IEEEauthorblockN{3\textsuperscript{rd} Tadashi Nakamura}
\IEEEauthorblockA{\textit{Chasis Engineering Dept. No.1} \\
\textit{Isuzu Motors Limited}\\
Tokyo, Japan \\
{Tadashi\_A\_Nakamura}@isuzu.com}
\and
\IEEEauthorblockN{4\textsuperscript{th} Hiroaki Sakata}
\IEEEauthorblockA{\textit{Chasis Engineering Dept. No.1} \\
\textit{Isuzu Motors Limited}\\
Tokyo, Japan \\
{Hiroaki\_Sakata}@isuzu.com}
\and
\IEEEauthorblockN{5\textsuperscript{th} Kazuya Seki}
\IEEEauthorblockA{\textit{Quality Assurance Dept.} \\
\textit{Isuzu Motors Limited}\\
Tokyo, Japan \\
{Kazuya\_Seki}@isuzu.com}
\and
\IEEEauthorblockN{6\textsuperscript{th} Yuu Teshigawara}
\IEEEauthorblockA{\textit{Quality Assurance Dept.} \\
\textit{Isuzu Motors Limited}\\
Tokyo, Japan \\
{Yuu\_Teshigawara}@isuzu.com} \\
\and
\IEEEauthorblockN{7\textsuperscript{th} Masami Yamashita}
\IEEEauthorblockA{\textit{Technology Strategy Dept.} \\
\textit{Isuzu Motors Limited}\\
Tokyo, Japan \\
{Masami\_Yamashita}@isuzu.com}
\and
\IEEEauthorblockN{8\textsuperscript{th} Kazuhiro Aoyama}
\IEEEauthorblockA{\textit{Department of System Innovation} \\
\textit{School of Engineering} \\
\textit{The University of Tokyo}\\
Tokyo, Japan \\
{aoyama}@race.t.u-tokyo.ac.jp}
}

\maketitle

\begin{abstract}
This paper presents a knowledge management system for automobile failure analysis using retrieval-augmented generation (RAG) with large language models (LLMs) and knowledge graphs (KGs). 
In the automotive industry, there is a growing demand for knowledge transfer of failure analysis from experienced engineers to young engineers. 
However, failure events are phenomena that occur in a chain reaction, making them difficult for beginners to analyze them. 
While knowledge graphs, which can describe semantic relationships and structure information is effective in representing failure events,
due to their capability of representing the relationships between components, 
there is much information in KGs, so it is challenging for young engineers to extract and understand sub-graphs from the KG. 
On the other hand, there is increasing interest in the use of Graph RAG, a type of RAG that combines LLMs and KGs for knowledge management. 
However, when using the current Graph RAG framework with an existing knowledge graph for automobile failures, 
several issues arise because it is difficult to generate executable queries for a knowledge graph database which is not constructed by LLMs. 
To address this, we focused on optimizing the Graph RAG pipeline for existing knowledge graphs. 
Using an original Q\&A dataset, the ROUGE F1 score of the sentences generated by the proposed method showed an average improvement of 157.6\% compared to the current method. 
This highlights the effectiveness of the proposed method for automobile failure analysis.
\end{abstract}

\begin{IEEEkeywords}
Graph RAG, Large Language Model, Knowledge Graph, Knowledge Management, Automobile Failure
\end{IEEEkeywords}

\section{Introduction}
In the Japanese automotive industry, especially truck industry, the necessity of knowledge transfer of failure analysis has been increasing in recent years, 
specifically from experienced engineers to young engineers. 
There are two main reasons for this.
The first reason is that failure analysis in automobiles is complex and challenging to address without sufficient experience.
Vehicles are generally recognized as systems composed of multiple parts, 
meaning that a failure in one part can potentially lead to a chain of failures in other components.
Therefore, understanding relationships of parts is crucial to find out the causal of failure, 
but this knowledge is mainly gained by experience not by lectures or textbooks.

The second one is that compared to the past, 
young engineers now need to address a broader and more specialized range of failure issues.
Japan is currently facing to a problem of declining birthrate and an aging population, resulting in a reduced number of young engineers compared to the past.
Moreover, there is growing complexity of automobile technologies owing to innovations such as CASE (Connected, Automated, Shared, Electric), 
Mobility as a Service (MaaS), and Electric Vehicles, so the burden on each young engineer has increased.

To address these challenges, a truck company issues and archives "\textbf{\textit{failure documents}}"
each time a failure occurs to ensure the transfer of failure analysis expertise.
These documents contain details about the conditions of malfunctioning vehicles, repair methods, dates, and circumstances, 
each organized in separate columns and written in Japanese. 
For effective knowledge transfer, it is necessary for young engineers to understand these documents. 
However, because they are written in natural language, it is difficult to comprehend the relationships between critical components in failure analysis.

To facilitate understanding of the connections between components, 
it is reasonable to represent these documents in the form of a knowledge graph (KG), 
which can describe semantic relationships and store structured information and is used for tackling problems in various domains\cite{ABUSALIH2021103076}. 
Hara et al. \cite{hara} has constructed \textbf{\textit{failure KG}} from \textit{failure documents},
but due to the vast amount of data within \textit{failure documents}, 
\textit{failure KG} also becomes complex and challenging for young engineers to comprehend.

Currently, there are some experienced engineers who can easily search the necessary information from the documents,
and they can provide the information to young engineers.
However, the average age of these engineers is not low, 
and the numbers of them are expected to decline in the future.
Additionally, truck failures cause substantial losses to both Japanese freights transportation, 
where trucks transport 91.4\% of the total weight of goods moved\footnote{\url{https://jta.or.jp/wp-content/themes/jta_theme/pdf/yusosangyo2023.pdfs}}, 
and truck drivers, so the company have to deal with the failures quickly.
Therefore, anticipating the retirement of experienced engineers, 
it is essential to develop a system that can quickly identify the causes of failures without relying on their experience. 
Specifically, it is desirable to construct a system that can critically present the causes when the fault conditions are inputted.

Large Language Models (LLMs), like ChatGPT\cite{ChatGPT}, demonstrate remarkable capabilities in various natural language processing tasks\cite{llm-kg},
and can response quickly, so they are expected to be effective for knowledge management of failure analysis.
However, since a vanilla LLM does not retain domain- or organization-specific information, 
an approach called Retrieval-Augmented Generation (RAG) is needed. 
This method answers user questions based on external information resources,
and Graph RAG\cite{edge2024localglobalgraphrag}, which is a type of RAG that combines LLMs and KGs, 
enables responses to users' queries based on the information within those graphs. 
However, current implementation of Graph RAG faces challenges in adapting to existing KGs.

In this study, we propose a novel Graph RAG system that can be applied to existing KGs. 
In the following section, we introduce related work, methodology—including current challenges—and the experiment using ROUGE\cite{lin-2004-rouge}. 
Main contributions of this research are summarized as follows.
\begin{itemize}
\item We propose a Graph RAG system that is independent of the query representation in knowledge graph databases.
\item The text generated by the proposed method achieved a higher ROUGE F1 score compared to those generated by the current Graph RAG and ChatGPT, 
demonstrating its effectiveness.
\end{itemize}

\section{Related Work}
\subsection{\textit{Failure Knowledge Graph}}\label{failure KG description}
\begin{figure}[htbp]
\centerline{\includegraphics[width=\linewidth]{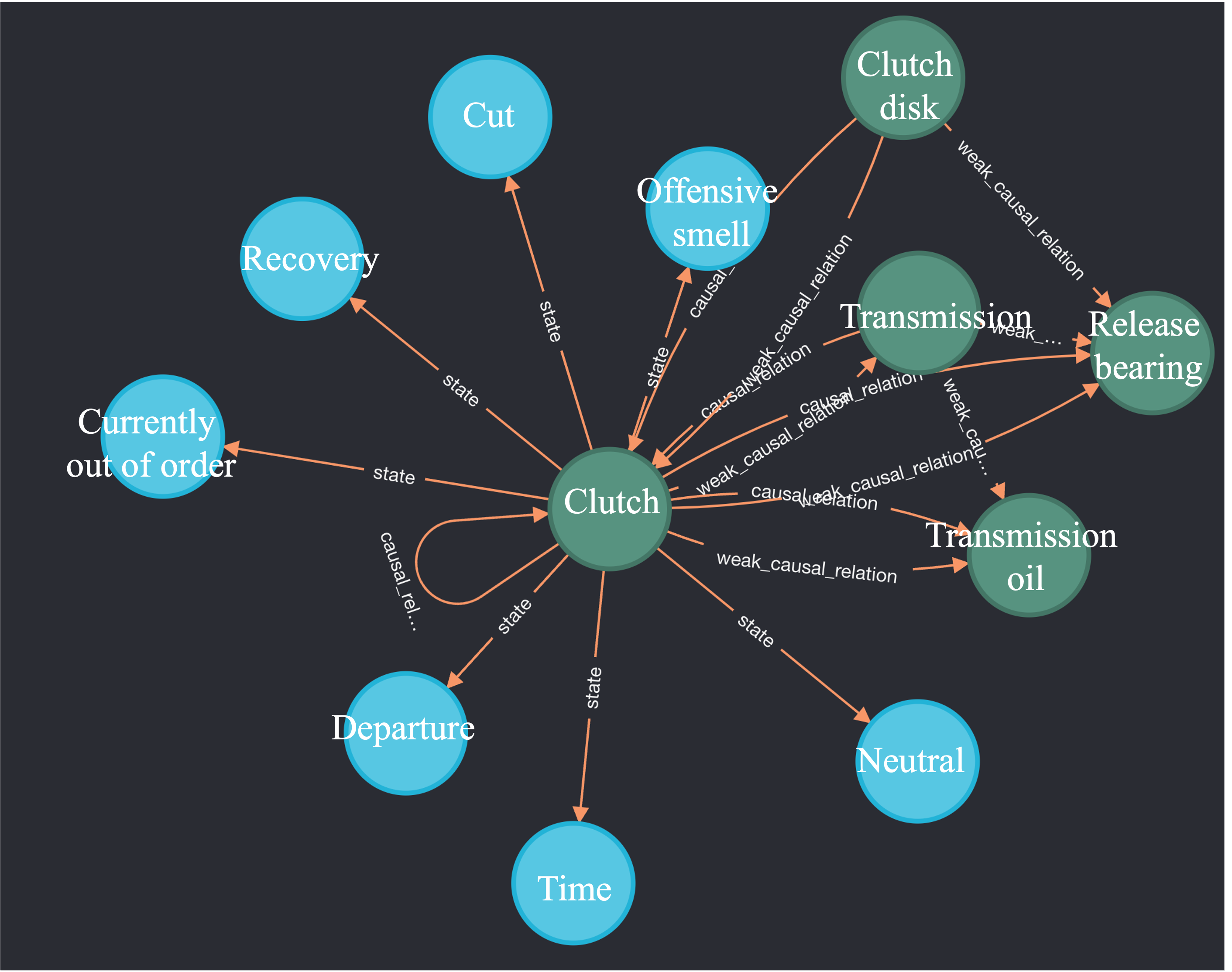}}
\caption{\textit{Failure KG}\cite{hara}. This figure illustrates the nodes related to the clutch node.
The original graph was created in Japanese, and the displayed excerpts are translated into English. 
In this representation, the blue labels represent the states of the system, while green labels denote the components involved.
}
\label{failure KG}
\end{figure}

Hara et al. \cite{hara} employed text mining techniques based on syntactic analysis, including the extraction of co-occurrence and dependency relations, 
as well as causal relation extraction using the algorithm proposed by Sakaji et al\cite{sakajiclue}. 
They constructed \textit{failure KG}, a knowledge graph of failure information. 
A portion of \textit{failure KG} is shown in Fig.~\ref{failure KG}.

Each node is labeled as a system, component, part, status, or other categories by experienced engineers. 
The meaning of each node is validated through human involvement. 
The relationships obtained from each syntactic analysis are categorized as weak causal and status relations for co-occurrence, 
hierarchical relations for dependencies, and causal relations for causal relation extraction.

While they demonstrated the ability to extract the interrelations of parts involved in a failure, 
challenges remain in understanding the resulting network especially for young engineers, due to vast amount of data.
Therefore, it is essential to implement the constructed knowledge graph in an application 
that can extract important sub-graphs from the KGs and generate answers for the question from young engineers.

\subsection{Large Language Model and RAG}
The release of ChatGPT by OpenAI\cite{ChatGPT} has demonstrated the vast potential of LLMs to generate sentences. 
These models can engage in continuous conversations with users based on chat context and perform complex tasks such as coding and academic writing\cite{llm-kg}. 
While LLMs can handle a wide range of questions, making them valuable for knowledge management, 
they often lack company-specific knowledge and may occasionally output incorrect information.
Consequently, there is a growing need for the development of LLMs tailored for specific companies.
However, unlike previous encoder-only language models like BERT\cite{Devlin2019BERTPO}, 
pre-training LLMs with specific domain knowledge or up-to-date information has become impractical owing to the significant amount of data, 
training time, and computational resources required, leading to extremely high costs. 
Additionally, contemporary LLMs, like ChatGPT, are generally offered solely as APIs, rendering pre-training unfeasible.

\begin{figure}[htbp]
\centerline{\includegraphics[width=\linewidth]{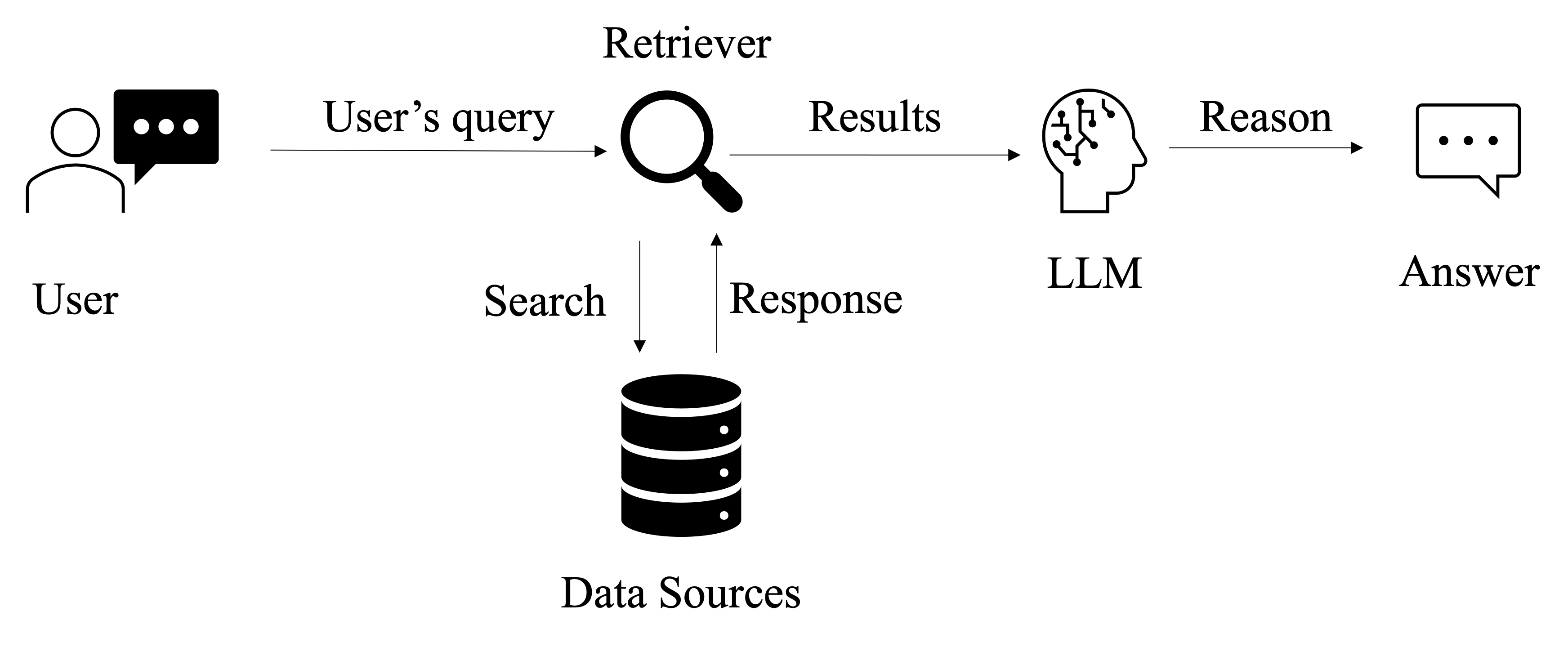}}
\caption{Overview of RAG}
\label{RAG}
\end{figure}

As a result, attention has shifted to Retriever-Augmented Generation (RAG) techniques.
Fig.~\ref{RAG} provides an overview of RAG.
This method involves retrieving necessary information from external data sources, such as document databases, 
based on the user's prompt and integrating this relevant information into the LLM's prompt. 
This effectively allows the LLM to incorporate the latest knowledge from specific domains\cite{gao2024retrievalaugmentedgenerationlargelanguage}. 
When using LLMs offered as APIs, RAG enables the integration of information without being limited by the amount of data held, 
while also reducing computational resource requirements. 

\subsection{Graph RAG}
\begin{figure}[htbp]
\centerline{\includegraphics[width=\linewidth]{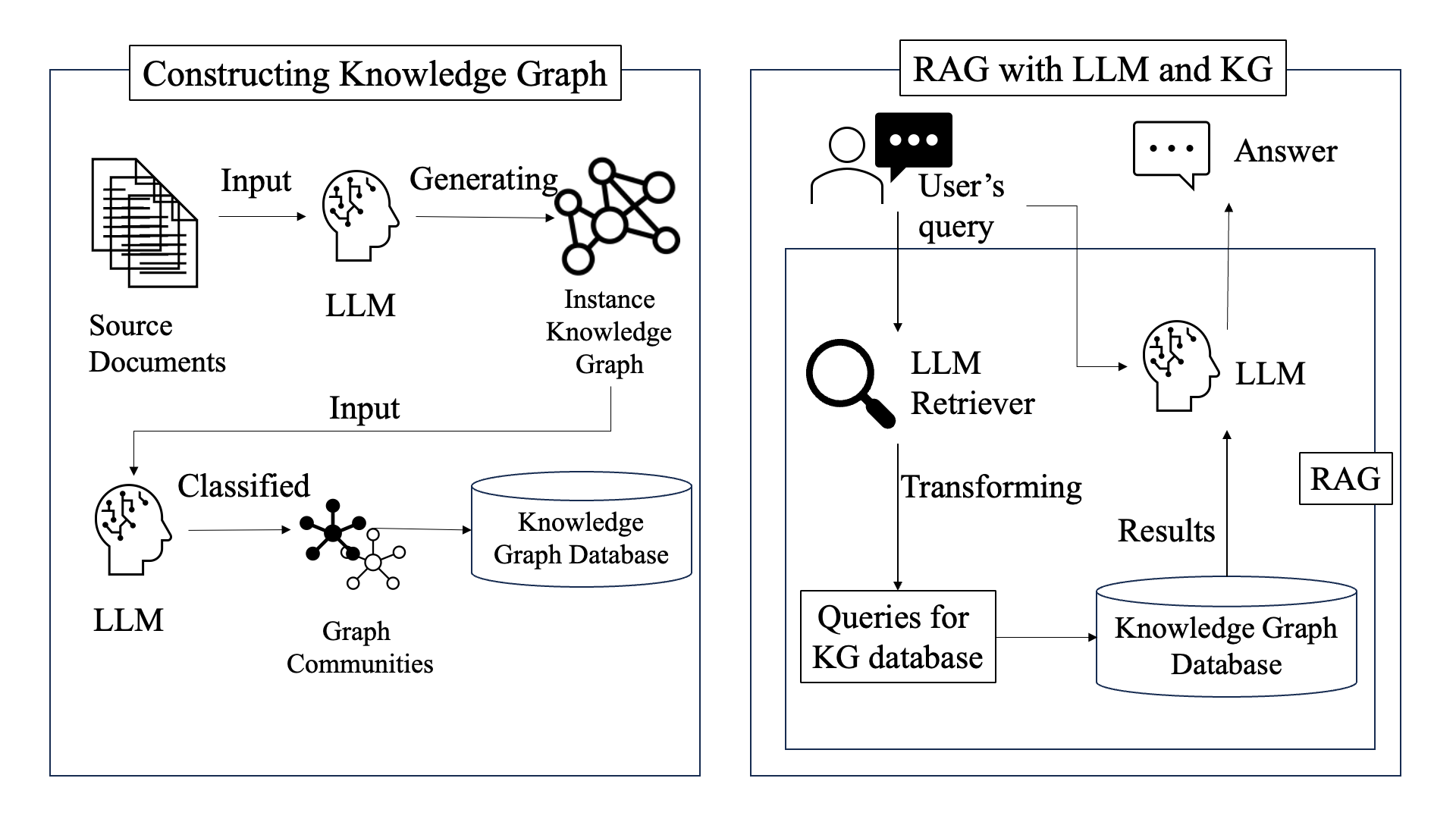}}
\caption{Overview of the Graph RAG}
\label{Graph RAG}
\end{figure}

Microsoft has proposed a Graph RAG system as a RAG framework that utilizes knowledge graphs \cite{edge2024localglobalgraphrag}. 
As Fig.~\ref{Graph RAG} provides, this system consists of two steps.

\begin{enumerate}
    \item \textit{Constructing the Knowledge Graph}: Source documents containing the information to be provided to the LLM are converted into a KG using the LLM and prompts. 
    After constructing the KG, the LLM forms graph communities, or clusters, 
    from the entire graph based on the number of relationships derived from the document. 
    This process can be viewed as labeling each node.
    \item \textit{RAG with LLM and KG}: First, the Retriever LLM generates a database query to extract information from the knowledge graph's database based on the user's query. 
    The LLM then generates an answer using the retrieved response and the user's query.
\end{enumerate}

Graph RAG can be viewed as executing a task categorized into Knowledge Graph-based Question Answering (KGQA), 
which focuses on providing answers to user-posed questions based on the KGs\cite{KGQA}. 
There are two main approaches to KGQA: the \textbf{Semantic Parsing} (SP) and \textbf{Information Retrieval} (IR) based methods\cite{KGQA}.
The SP method involves converting natural language questions posed by users into search queries for the knowledge graph, 
specifically into logical symbols, and then generating answers based on the results obtained from these queries\cite{KGQA}. 
On the other hand, the IR method involves extracting an understandable sub-graph from the knowledge graph that is related to the question, 
reasoning based on this sub-graph, and generating the answer\cite{KGQA}. 

In this process, the user's question is converted into a searchable query by the LLM, which is then sent to the knowledge graph database. 
The LLM generates an answer based on the information retrieved from the knowledge graph database.

Additionally, the combination of KGs and LLMs has been reported to generate more consistent answers and suppress the generation of misinformation, 
commonly referred to as "hallucinations", compared to other RAG approaches\cite{gao2024retrievalaugmentedgenerationlargelanguage}.

\section{Methodology}
\subsection{Problem Definition}\
When applying the current Graph RAG to automotive fault knowledge management, two major challenges emerge. 
The first challenge is the difficulty in adapting to existing knowledge graphs. 
As previously mentioned, the current Graph RAG is based on the SP method. 
It can be considered that executable queries can be generated without high-cost training, which is generally needed\cite{KGQA},
as the same LLM is utilized both for converting a user query $q$ from the natural language space to a query $q$ for KG databases in logical symbol space, 
as well as for constructing KGs from documents. 
Therefore, while Graph RAG is effective for KGs constructed by LLM, it remains unclear whether it can be applied to existing KGs created using other methods. 
As stated in \ref{failure KG description}, \textit{failure KG} is difficult for beginners to comprehend, 
but it has been confirmed that the relationships between components during a failure can be partially extracted. 
Therefore, in this study, we want to focus on utilizing this KG as it currently exists.
Additionally, the knowledge graphs in Graph RAG often lack human-validated meaning, making them insufficient for practical use.

The second challenge is that the relationships and structures described in the knowledge graph are rarely used in answer generation. 
Since Graph RAG operates using the SP method, the search typically targets the graph's nodes rather than its edges, i.e., the relationships. 
However, automobile failures involve chains and their structures, making it essential to use the information from entire sub-graphs containing the necessary information, 
rather than focusing on specific nodes.

These challenges primarily arise from the fact that the current Graph RAG constructs a pipeline based on the SP method. 
In response, we propose an \textbf{IR-based Graph RAG} that is applicable to existing knowledge graphs and is based on the IR method. 
While it has been mentioned that the IR method may lead to an increase in the number of extracted sub-graphs in response to complex queries, 
there is potential to resolve this issue using LLMs. 

\subsection{Proposed Method}
\begin{figure}[htbp]
\centerline{\includegraphics[width=\linewidth]{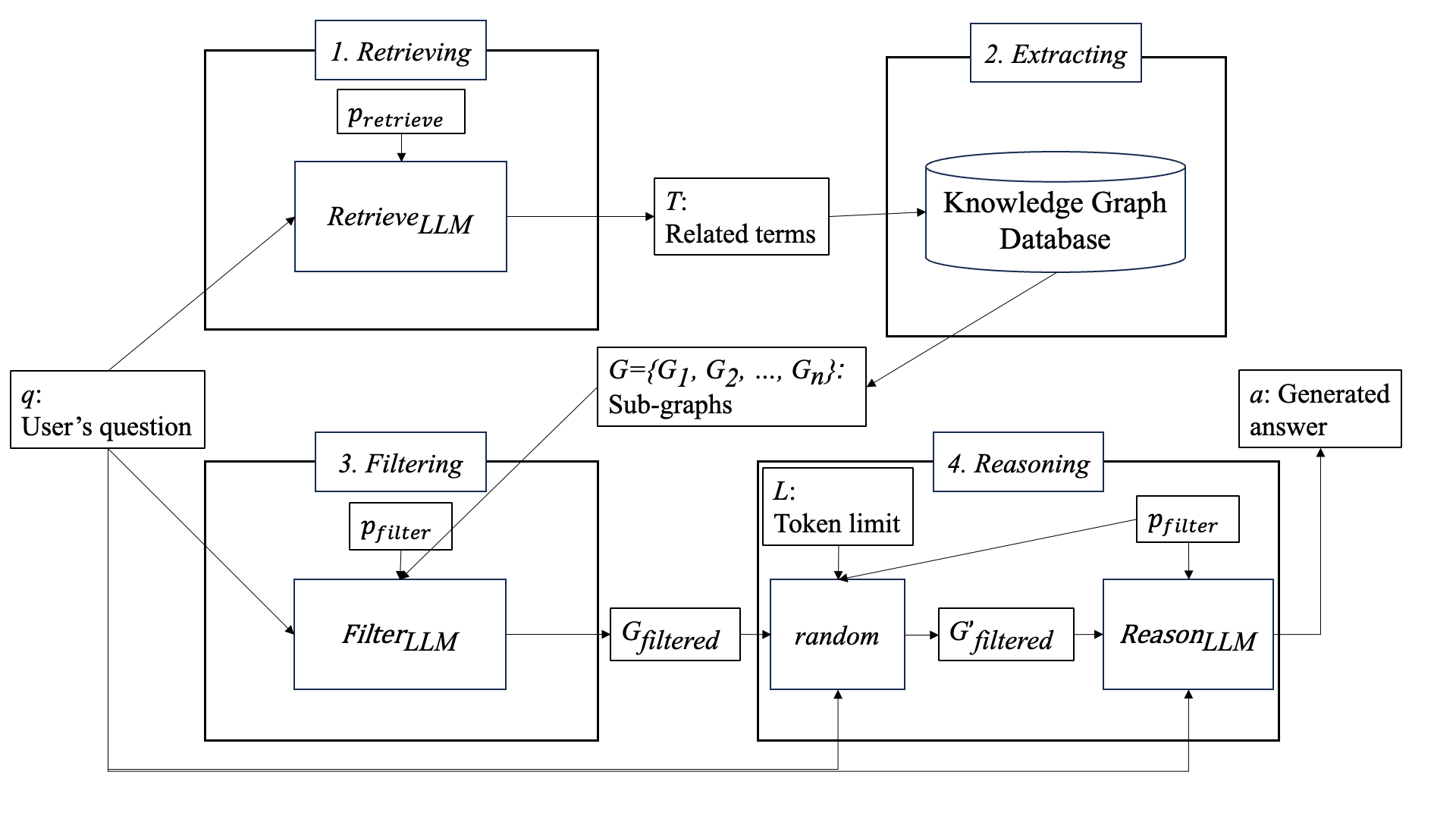}}
\caption{Overview of the IR-based Graph RAG}
\label{Proposed Method}
\end{figure}

The following outlines the pipeline of the IR-based Graph RAG. 
A schematic diagram is presented in Fig.~\ref{Proposed Method}.

\begin{figure}[htbp]
\centerline{\includegraphics[width=\linewidth]{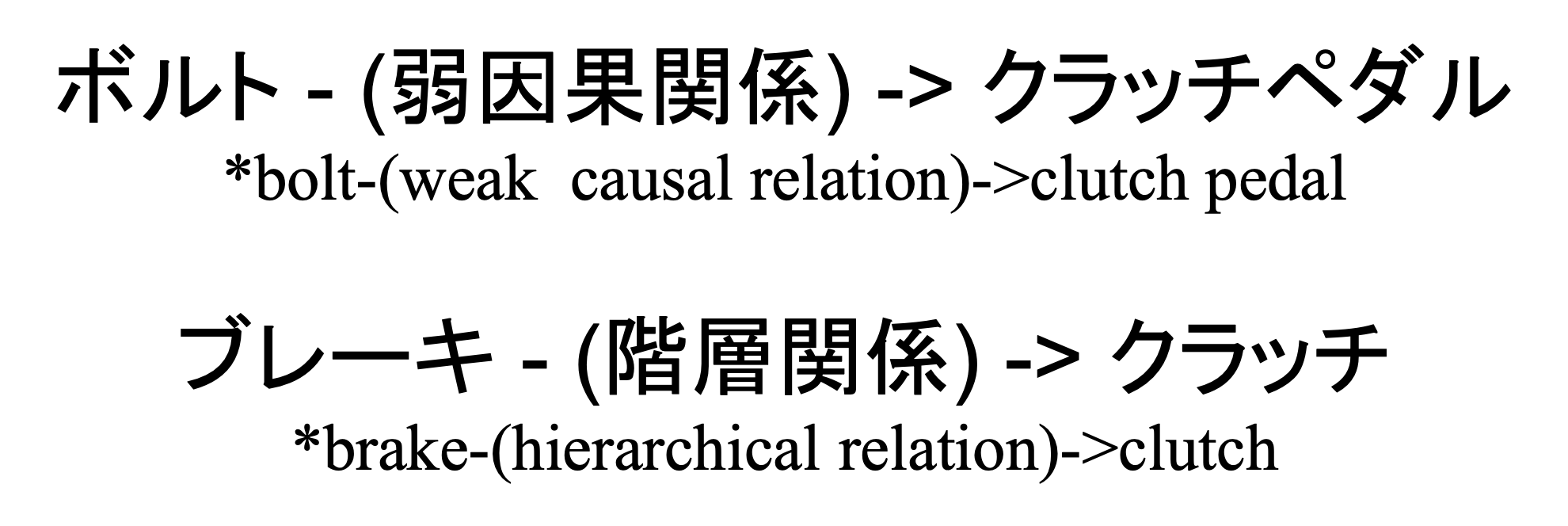}}
\caption{Examples of sub-graph representations}
\label{sub-graph-img}
\end{figure}

\begin{enumerate}
    \item \textit{Retrieving}: Given a user query $q$, the LLM processes the input and retrieves a set of related terms $T$ by identifying relevant automotive issues and faults. 
    Mathematically, this can be expressed as below:
    \begin{equation}
    T=Retrieve_{LLM}(q,p_{retrieve})\label{retrieve}
    \end{equation}
    where $p_{retrieve}$ is the prompt instructing the retrieval of related words.
    \item \textit{Extracting}: Sub-graphs are extracted based on the retrieved words. 
    This extraction involves generating a rule-based search query that is free from grammatical errors. 
    For each word $t \in T$ , a sub-graph $G_t$  is defined as s set of relationships, including the node $n_t$ corresponding to the target word $t$, 
    along with the edges $E_t$  and connected nodes $N_t$. 
    Mathematically, the sub-graph $G_t$  can be expressed as:
    \begin{equation}
    G_t=(n_t, N_t, E_t )\label{sub-graph}
    \end{equation}
    where $n_t$ is the target word node, $N_t$  is the set of nodes connected to $n_t$, 
    and $E_t$  is the set of edges connecting $n_t$  to $N_t$, or $N_t$  to $n_t$. 
    In other words, these relationships form one-hop chain. 
    Also, the sub-graph $G_t$ is then represented as sets of text shown in Fig.~\ref{sub-graph-img}.
    \item \textit{Filtering}:  The LLM is used to filter the extracted sub-graphs, selecting candidates that are relevant to the user's question. 
    This step is necessary because many edges are connected to specific words (such as “engine”), resulting in the extraction of numerous unrelated sub-graphs.
    By using the LLM, this step mitigates the conventional issue of an overwhelming number of extracted sub-graphs when adopting the IR-method for KGQA. 
    Let $G={G_1, G_2,\ldots ,G_n}$ represent the set of extracted sub-graphs. 
    The LLM applies a filtering function $Filter_{LLM}$ that selects a subset $G_{filtered} \subseteq G$ of sub-graphs relevant to the query $q$, 
    which can be expressed as:
    \begin{equation}
    G_{filtered}=Filter_{LLM}(G,q,p_{filter})\label{filtering}
    \end{equation}
    where $p_{filter}$ is the prompt containing instructions to filter the sub-graphs.
    \item \textit{Reasoning}: The final filtered sub-graphs, along with the user's question, are given to the LLM as a prompt to generate the final answer. 
    If the number of selected sub-graphs is too large and exceeds the token limit $L$ for the prompt, 
    random sub-graphs are removed before included in the prompt. 
    This can be expressed as:
    \begin{equation}
    G'_{filtered}=random(G_{filtered},L,q,p_{reason})\label{random}
    \end{equation}
    where $p_{reason}$ is the prompt used to instruct the LLM to reason and generate the answer.
    The final answer $a$ is then generated by the LLM as expressed below:
    \begin{equation}
    a=Reason_{LLM}(G'_{filtered},q,p_{reason})\label{reasoning}
    \end{equation}
\end{enumerate}

\subsection{Evaluation}
To evaluate the proposed method, we assess its validity by creating and using a Q\&A dataset to measure the textual similarity between the responses generated by Graph RAG and the expected answers. 
The dataset is generated by providing the LLM with \textit{failure documents} and specific instructions as prompts, resulting in a set of questions and answers related to malfunction or failure information. 
To make the content more practical, the dataset is designed so that the answers involve the actual component configurations and 
failure propagation processes typically analyzed during failure analysis.
The sentences from \textit{failure documents} are cleansed and converted into grammatically correct and meaningful sentences by the LLM.
This approach is similar to that of Balaguer et al\cite{balaguer2024ragvsfinetuningpipelines}.

For the evaluation metric, we use the ROUGE F1 score \cite{lin-2004-rouge}. 
We hypothesize that when questions generated from \textit{failure documents} and the LLM are input into Graph RAG, which uses \textit{failure KG} as the data source, 
the responses will be similar to those generated by \textit{failure documents} and LLM. 
A higher ROUGE score indicates greater textual similarity with the dataset, suggesting that the generated response contains more of the required information. 
Therefore, we adopted this metric for this study.

\section{Experiment}
\subsection{Experiment Setting}
In this study, we exclusively used ChatGPT as the LLM. 
Specifically, we employed the Azure OpenAI GPT-4o (2024-07-01) with the temperature parameter set to 0.

We compared three methods: the proposed method, ChatGPT, and the current Graph RAG. 
Both versions of Graph RAG use \textit{failure KG}. 
Because the current Graph RAG adopts the SP method, we named it \textbf{SP-based Graph RAG}.
Additionally, to implement SP-based Graph RAG, we used libraries supported by LangChain and Neo4j\footnote{\url{https://python.langchain.com/v0.2/docs/integrations/graphs/neo4j_cypher/}}.

In the prompt for the SP-based Graph RAG, we include not only the reasoning prompt used in the IR-based Graph RAG but also an explanatory description of \textit{failure KG}.

The SP-based Graph RAG was executed once, while the other two methods were run five times each to calculate the average score. 
This approach was chosen because of the variability in the generated text; 
when the token limit was exceeded, relationships are randomly deleted, leading to fluctuations in the scores. 

For the evaluation, we used the F1 scores of ROUGE-1, ROUGE-2, and ROUGE-L. 
We computed the ROUGE F1 scores for each set in the dataset, summed them, and then divided the total by the number of sets.

\subsection{Dataset}
\textit{failure documents} were provided by Isuzu Motors Limited. 
For the dataset, we used 43 sets of \textit{failure documents} that included the term \begin{CJK}{UTF8}{min}クラッチ\end{CJK} (clutch) and contained all the necessary information for constructing \textit{failure KG}. 
We instructed the LLM to create a dataset of questions and answers related to the chain of parts or failures.
We used GPT-4 (1106-preview) provided by Azure OpenAI to construct the dataset, setting the temperature parameter to 0.

\subsection{Results}
The results are presented in Table~\ref{tab1}. 
For the SP-based Graph RAG, five sets of the dataset encountered errors and were excluded from the evaluation. 
All errors were attributed to syntax issues in the search queries generated for the KG, with successful query generation achieved for only ten sets.

The IR-based Graph RAG demonstrated an average score improvement of 157.6\% compared to the current Graph RAG and an average improvement of 23.18\% over ChatGPT. 
The IR-based Graph RAG consistently achieved the highest scores in all cases, indicating that it contained the most relevant information for the answers. 
Conversely, the SP-based Graph RAG produced very low scores, challenging to consider it a practical text. 
Furthermore, generating effective search queries proved to be challenging. 
Even when queries were successfully created, a significant number of syntax errors occurred, 
suggesting that this method may not be suitable for use with existing knowledge graphs.

\begin{table}[htbp]
\caption{The Experiment Results.}
\begin{center}
\begin{tabular}{|c|c|c|c|}
\hline
\textbf{Method} & \textbf{ROUGE-1 F1}& \textbf{ROUGE-2 F1}& \textbf{ROUGE-L F1} \\
\hline
ChatGPT& 0.3305 & 0.1483 & 0.2364 \\
\hline
SP-based& 0.1538 & 0.06967 & 0.1185 \\
\hline
IR-based& \textbf{0.4053} & \textbf{0.1845} & \textbf{0.2896} \\
\hline
\end{tabular}
\label{tab1}
\end{center}
\end{table}

Additionally, the proposed method outperformed ChatGPT, 
clearly demonstrating its capability to extract essential information from the \textit{failure KG}.
This indicates that the proposed approach is more effective for automotive failure knowledge management compared to existing technologies.
Furthermore, when comparing the average number of tokens in the text generated by ChatGPT and the IR-based Graph RAG, 
the latter produces fewer tokens (as shown in Table~\ref{tokens}), suggesting that it generates more concise text.
This may make it easier for beginners to understand.

\begin{table}[htbp]
\caption{The Average Tokens.}
\begin{center}
\begin{tabular}{|c|c|c|c|}
\hline
\textbf{Method} & \textbf{Average Tokens}\\
\hline
ChatGPT& 530.2 \\
\hline
IR-based& 320.0 \\
\hline
\end{tabular}
\label{tokens}
\end{center}
\end{table}

\section{Discussion}
\subsection{Cause Analysis about Missing Information}
While the proposed method has demonstrated effectiveness, the evaluation metric scores remain low, raising concerns about its practicality. 
Possible reasons for this include missing information that should be stored in \textit{failure KG} and the possibility that the method for extracting and describing sub-graphs is not appropriate. 
In this study, we conduct additional verification of the former issue.

To demonstrate that the amount of information recorded in \textit{failure KG} is insufficient, 
we propose comparing the generated answers by adding sentences from \textit{failure documents} that formed the basis of \textit{failure KG} into the prompts. 
We hypothesize that including these sentences in the Reasoning step will improve the evaluation metrics, 
as they are expected to contain more information.
Specifically, we add an evaluation method called IR-based Graph RAG with sentences (mentioned as "With sentences" in Tables), where sentences from \textit{failure documents} 
that generated the sub-graph are also extracted in the Filtering step and added as prompts alongside the sub-graphs during the Reasoning step. 
Thus, the following equation is added to the Filtering step:
\begin{equation}
S=Extract(G_{filtered})\label{sentence_extract}
\end{equation}
where $S$ is the set of sentences extracted from \textit{failure documents} that generated the sets of sub-graphs $G_{filtered}$.

Also, during the Reasoning step, these processes are executed:
\begin{equation}
G'_{filtered},S'=random(G_{filtered},S,L,q,p_{reason})\label{sentence_with_random}
\end{equation}
\begin{equation}
a_{with~sentences}=Reason_{LLM}(G'_{filtered},S',q,p_{reason})\label{sentence_with_reaso}
\end{equation}
where $S'$ are randomly selected sentences to avoid exceed the token limit $L$, and $a_{with~sentences}$  is the generated answer by this method, which was used for evaluation.

Additionally, we include a separate method called IR-based Graph RAG only sentences (mentioned as "Only sentences" in Tables), 
in which sentences from \textit{failure documents} are extracted during the Extracting step, similar to the IR-based Graph RAG with documents.
However, in the Reasoning step, only these sentences are given as prompts. 
Therefore, adding to the pipeline of IR-based Graph RAG, \eqref{sentence_extract} is executed during the Filtering step, 
and the following processes are executed in the Reasoning step:
\begin{equation}
S'=random(S,L,q,p_{reason})\label{sentence_only_random}
\end{equation}
\begin{equation}
a_{only~sentences}=Reason_{LLM}(G'_{filtered},S',q,p_{reason})\label{sentence_only_reason}
\end{equation}
where $a_{only~sentences}$  is the generated answer used for evaluation.

Each method was run five times, and the average ROUGE F1 score calculated for each run was used. 
The other experimental conditions remain the same as in the previously conducted experiments. 
The results are shown in Table~\ref{tab2}.

Based on the results, including both the sub-graphs and sentences from \textit{failure documents} in the prompt during the Reasoning step leads to more accurate output. 
The scores of IR-based Graph RAG with sentences were the highest, 
showing an average improvement of 8.18\% compared to the proposed method, which can be recognized as a significant difference. 
This suggests that the extracted sub-graph alone is insufficient and that there is missing information that should be described in \textit{failure KG}. 

\begin{table}[htbp]
\caption{Results of The Experiment about Missing Information.}
\begin{center}
\begin{tabular}{|c|c|c|c|}
\hline
\textbf{Method} & \textbf{ROUGE-1 F1}& \textbf{ROUGE-2 F1}& \textbf{ROUGE-L F1} \\
\hline
With sentences& 0.4245 & 0.2106 & 0.3060 \\
\hline
Only sentences& 0.3940 & 0.1891 & 0.2825 \\
\hline
\end{tabular}
\label{tab2}
\end{center}
\end{table}

\begin{figure}[htbp]
\centerline{\includegraphics[width=\linewidth]{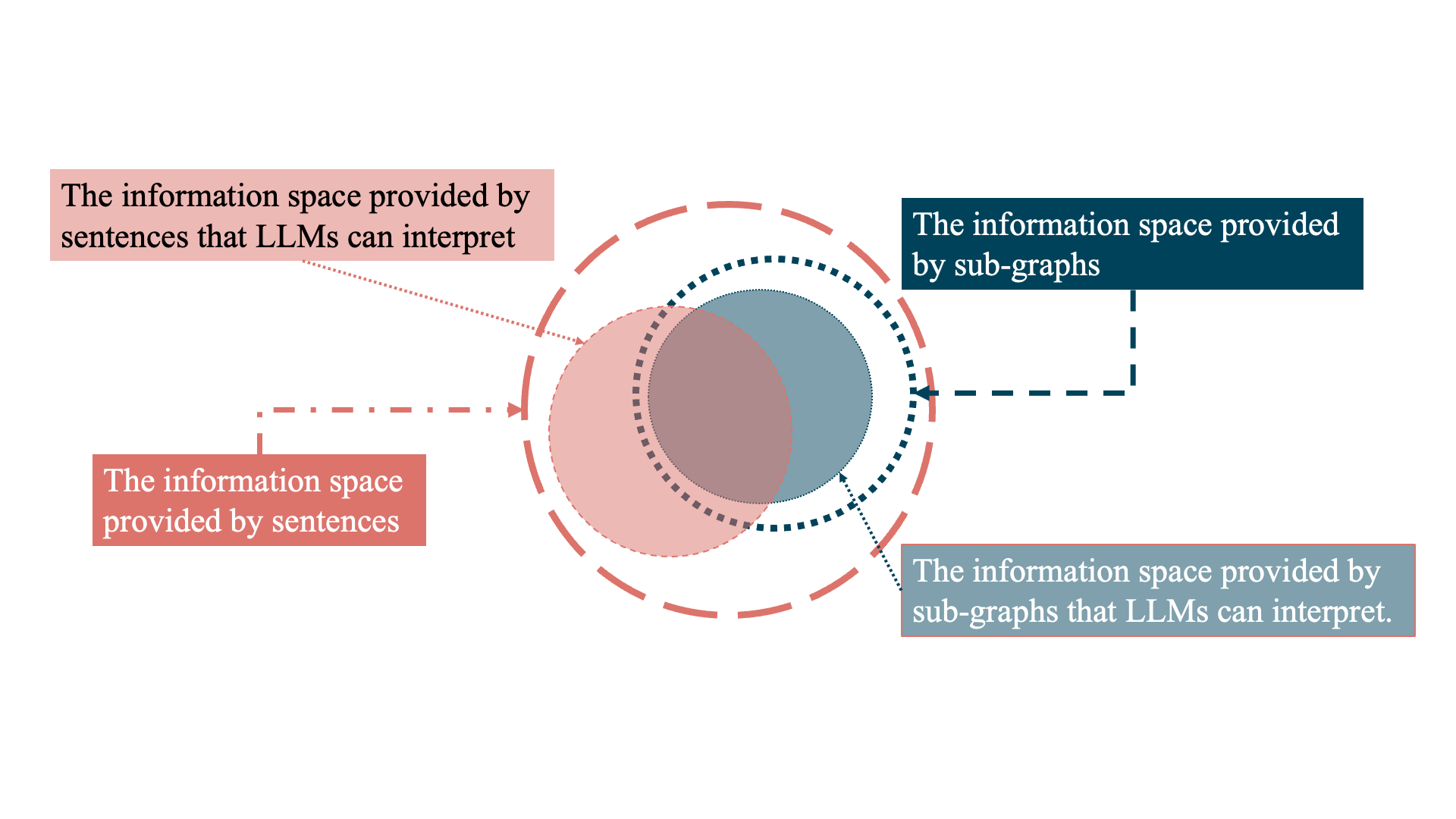}}
\caption{Comparison of the information space provided by sentences versus sub-graphs}
\label{information_space}
\end{figure}

While the additional experiment demonstrated that there was missing information that should have been included in \textit{failure KG}.
However, there was no significant difference observed between the IR-based Graph RAG and the IR-based Graph RAG only sentences. 
This can be explained by the proportion of information that LLMs can interpret. 
Generally, text of \textit{failure documents} is considered to provide more information as a prompt compared to the sub-graphs. 
However, because \textit{failure documents} are written in natural language, they likely contain a lot of information that could be considered noise. 
Therefore, the proportion of interpretable information from \textit{failure documents} is likely smaller compared to that of the sub-graphs by the LLM. 
Additionally, the difference in the distribution of interpretable information between the sub-graphs and \textit{failure documents} suggests that providing both sources of information in the prompt results in the highest score.
In summary, these explanations can be represented as shown in Fig.~\ref{information_space}.

\subsection{Ablation Analysis about the Filtering step}
To verify the effectiveness of the Filtering step in the proposed method, we conducted an ablation analysis.
In this analysis, we removed the Filtering step from the pipelines and evaluated the ROUGE F1 score.
We compare the vanilla IR-based Graph RAG, IR-based Graph RAG only sentences, and IR-based Graph RAG with sentences,
and use the same experimental conditions as in the previous experiment, and the results are shown in Table~\ref{tab3}.

\begin{table}[htbp]
\caption{The Ablation Analysis Results of the IR-based Graph RAG.}
\begin{center}
\begin{tabular}{|c|c|c|c|}
\hline
\textbf{Method} & \textbf{ROUGE-1 F1}& \textbf{ROUGE-2 F1}& \textbf{ROUGE-L F1} \\
\hline
Vanilla& 0.3882 & 0.1769 & 0.2809 \\
\hline
Only sentences& 0.3695 & 0.1713 & 0.2632 \\
\hline
With sentences& 0.3827 & 0.1885 & 0.2768 \\
\hline
\end{tabular}
\label{tab3}
\end{center}
\end{table}

The results garnered indicate an average diminution of 7.077\% relative to preceding scores, 
underscoring the indispensability of the filtering phase within the proposed methodology. 
This implies that the filtering stage in the proposed approach is not merely a procedural necessity, 
but a critical component, given that the selection of pertinent sub-graphs profoundly influences the precision of the outcomes.
Additionally, it was found that filtering sub-graphs, which had been pointed out as a challenge in IR methods\cite{KGQA}, 
can be performed on LLMs without requiring any special training.

\subsection{Limitation and future works}
This study has three major limitations. 
The first is the inadequacy of the sub-graph extraction method. 
In the context of failure events, chains of events are crucial, but this study extracts only one-hop chain. 
Future work should explore the introduction of an algorithm that can extract multi-hop chains.
For example, it is a possible approach that embedding nodes or edges using an encoder model like BERT\cite{Devlin2019BERTPO},
and then determining which chains in KGs to use for reasoning based on cosine similarity to the query.
This approach is similar to the method proposed by Saxena et al\cite{saxena-etal-2020-improving}.

The second limitation concerns the evaluation method. 
In this study, the dataset was constructed using ChatGPT, and the evaluation was based on ROUGE F1. 
However, ROUGE is intended for use in scenarios where the target text is directly used, such as in extractive summarization, 
and its suitability for evaluating generated text remains questionable. 
To enhance the credibility of the evaluations, it is necessary to incorporate human evaluations as well.

The last one is that only a single round of interaction is conducted during the Reasoning step.
In the experiments, there are some responses that simply include the names of components without providing deeper explanations. 
This is likely due to the attempt to generate an answer within a single interaction.
Recent LLMs support agent systems, allowing them to act autonomously. 
These agents can analyze what information is needed to answer a given query, 
utilize provided tools (e.g., databases, web searches) to retrieve the required information, and perform reasoning based on it\cite{Wang2024}. 
Depending on the prompt, they can autonomously execute multi-step processes, 
which enables them to generate more accurate and contextually appropriate responses.

\section{Conclusion}
This study demonstrates that the proposed method, IR-based Graph RAG, is an effective system for knowledge management in automotive failure analysis. 
It suggests that providing information in the form of sub-graphs as prompts offers an interpretable format for LLMs. 
However, it is important to note that, that at this stage, it has not yet reached a level where it can be used directly, 
and we have identified issues specifically with the knowledge graph used in Graph RAG. 

In the future, we intend to verify the practicality and validity for a wider range of events by increasing the size of the dataset.
Additionally, we aim to improve the score accuracy through prompt engineering as part of our future work.

\section*{Acknowledgment}

This study was supported by Isuzu Motors Limited.

\bibliographystyle{IEEEtran}
\bibliography{IEEEabrv,citation}

\begin{thebibliography}{10}
\providecommand{\url}[1]{#1}
\csname url@samestyle\endcsname
\providecommand{\newblock}{\relax}
\providecommand{\bibinfo}[2]{#2}
\providecommand{\BIBentrySTDinterwordspacing}{\spaceskip=0pt\relax}
\providecommand{\BIBentryALTinterwordstretchfactor}{4}
\providecommand{\BIBentryALTinterwordspacing}{\spaceskip=\fontdimen2\font plus
\BIBentryALTinterwordstretchfactor\fontdimen3\font minus \fontdimen4\font\relax}
\providecommand{\BIBforeignlanguage}[2]{{%
\expandafter\ifx\csname l@#1\endcsname\relax
\typeout{** WARNING: IEEEtran.bst: No hyphenation pattern has been}%
\typeout{** loaded for the language `#1'. Using the pattern for}%
\typeout{** the default language instead.}%
\else
\language=\csname l@#1\endcsname
\fi
#2}}
\providecommand{\BIBdecl}{\relax}
\BIBdecl

\bibitem{ABUSALIH2021103076}
\BIBentryALTinterwordspacing
B.~Abu-Salih, ``Domain-specific knowledge graphs: A survey,'' \emph{Journal of Network and Computer Applications}, vol. 185, p. 103076, 2021. [Online]. Available: \url{https://www.sciencedirect.com/science/article/pii/S1084804521000990}
\BIBentrySTDinterwordspacing

\bibitem{hara}
S.~Hara, H.~Ozawa, M.~Yamashita, S.~Yamada, Y.~Sakachi, and K.~Aoyama, ``Knowledge management for design development and maintenance through text mining of product information (knowledge management for design development and maintenance using product development history information and product defect response information),'' \emph{Design \& Systems Conference}, vol. 2022.32, p. 1201, Jan. 2022.

\bibitem{ChatGPT}
L.~Ouyang, J.~Wu, X.~Jiang, D.~Almeida, C.~L. Wainwright, P.~Mishkin, C.~Zhang, S.~Agarwal, K.~Slama, A.~Ray, J.~Schulman, J.~Hilton, F.~Kelton, L.~Miller, M.~Simens, A.~Askell, P.~Welinder, P.~Christiano, J.~Leike, and R.~Lowe, ``Training language models to follow instructions with human feedback,'' in \emph{Proceedings of the 36th International Conference on Neural Information Processing Systems}, ser. NIPS '22.\hskip 1em plus 0.5em minus 0.4em\relax Red Hook, NY, USA: Curran Associates Inc., 2024.

\bibitem{llm-kg}
L.~Yang, H.~Chen, Z.~Li, X.~Ding, and X.~Wu, ``Give us the facts: Enhancing large language models with knowledge graphs for fact-aware language modeling,'' \emph{IEEE Transactions on Knowledge and Data Engineering}, vol.~PP, pp. 1--20, 07 2024.

\bibitem{edge2024localglobalgraphrag}
\BIBentryALTinterwordspacing
D.~Edge, H.~Trinh, N.~Cheng, J.~Bradley, A.~Chao, A.~Mody, S.~Truitt, and J.~Larson, ``From local to global: A graph rag approach to query-focused summarization,'' 2024. [Online]. Available: \url{https://arxiv.org/abs/2404.16130}
\BIBentrySTDinterwordspacing

\bibitem{lin-2004-rouge}
\BIBentryALTinterwordspacing
C.-Y. Lin, ``{ROUGE}: A package for automatic evaluation of summaries,'' in \emph{Text Summarization Branches Out}.\hskip 1em plus 0.5em minus 0.4em\relax Barcelona, Spain: Association for Computational Linguistics, Jul. 2004, pp. 74--81. [Online]. Available: \url{https://aclanthology.org/W04-1013}
\BIBentrySTDinterwordspacing

\bibitem{sakajiclue}
H.~Sakaji, S.~Sekine, and S.~Masuyama, ``Extracting causal knowledge using clue phrases and syntactic patterns,'' in \emph{Practical Aspects of Knowledge Management}, T.~Yamaguchi, Ed.\hskip 1em plus 0.5em minus 0.4em\relax Berlin, Heidelberg: Springer Berlin Heidelberg, 2008, pp. 111--122.

\bibitem{Devlin2019BERTPO}
\BIBentryALTinterwordspacing
J.~Devlin, M.-W. Chang, K.~Lee, and K.~Toutanova, ``Bert: Pre-training of deep bidirectional transformers for language understanding,'' in \emph{North American Chapter of the Association for Computational Linguistics}, 2019. [Online]. Available: \url{https://api.semanticscholar.org/CorpusID:52967399}
\BIBentrySTDinterwordspacing

\bibitem{gao2024retrievalaugmentedgenerationlargelanguage}
\BIBentryALTinterwordspacing
Y.~Gao, Y.~Xiong, X.~Gao, K.~Jia, J.~Pan, Y.~Bi, Y.~Dai, J.~Sun, M.~Wang, and H.~Wang, ``Retrieval-augmented generation for large language models: A survey,'' 2024. [Online]. Available: \url{https://arxiv.org/abs/2312.10997}
\BIBentrySTDinterwordspacing

\bibitem{KGQA}
Y.~Luo, B.~Yang, D.~Xu, and L.~Tian, ``A survey: Complex knowledge base question answering,'' in \emph{2022 IEEE 2nd International Conference on Information Communication and Software Engineering (ICICSE)}, 2022, pp. 46--52.

\bibitem{balaguer2024ragvsfinetuningpipelines}
\BIBentryALTinterwordspacing
A.~Balaguer, V.~Benara, R.~L. de~Freitas~Cunha, R.~de~M.~Estevão~Filho, T.~Hendry, D.~Holstein, J.~Marsman, N.~Mecklenburg, S.~Malvar, L.~O. Nunes, R.~Padilha, M.~Sharp, B.~Silva, S.~Sharma, V.~Aski, and R.~Chandra, ``Rag vs fine-tuning: Pipelines, tradeoffs, and a case study on agriculture,'' 2024. [Online]. Available: \url{https://arxiv.org/abs/2401.08406}
\BIBentrySTDinterwordspacing

\bibitem{saxena-etal-2020-improving}
\BIBentryALTinterwordspacing
A.~Saxena, A.~Tripathi, and P.~Talukdar, ``Improving multi-hop question answering over knowledge graphs using knowledge base embeddings,'' in \emph{Proceedings of the 58th Annual Meeting of the Association for Computational Linguistics}, D.~Jurafsky, J.~Chai, N.~Schluter, and J.~Tetreault, Eds.\hskip 1em plus 0.5em minus 0.4em\relax Online: Association for Computational Linguistics, Jul. 2020, pp. 4498--4507. [Online]. Available: \url{https://aclanthology.org/2020.acl-main.412}
\BIBentrySTDinterwordspacing

\bibitem{Wang2024}
\BIBentryALTinterwordspacing
L.~Wang, C.~Ma, X.~Feng, Z.~Zhang, H.~Yang, J.~Zhang, Z.~Chen, J.~Tang, X.~Chen, Y.~Lin, W.~X. Zhao, Z.~Wei, and J.~Wen, ``A survey on large language model based autonomous agents,'' \emph{Frontiers of Computer Science}, vol.~18, no.~6, p. 186345, Mar 2024. [Online]. Available: \url{https://doi.org/10.1007/s11704-024-40231-1}
\BIBentrySTDinterwordspacing

\end{thebibliography}

\end{document}